\documentclass[12pt]{article}
\usepackage{rotating,amsmath, amssymb, amsthm}
\usepackage{amsfonts, verbatim,url}
\usepackage{epsfig,url, psfrag,appendix}
\usepackage{graphicx, multirow,graphics}
\usepackage{setspace}
\usepackage[left=3.1cm,top=2.5cm,right=3.1cm,bottom=2.5cm]{geometry}
\usepackage{natbib}
\usepackage{float}
\floatstyle{ruled}

\newtheorem{thm}{Theorem}

\newtheorem{rem}{Remark}
\newcommand{\B}{\boldsymbol}
\newcommand{\M}{\mathbf}

\newcommand{\s}{\mathbf S}

\newcommand{\half}{\mbox{$\frac12$}}

\newcommand{\tr}{\,\mathrm{tr}}

\def\calE{\widehat{\B{\cal E}}}
\def\bfE{\B{\cal E}}
\def\calG{\widehat{\cal G}}
\def\bfG{{\cal G}}
\def\kaplam{\widehat{k(\lambda)}}

\def\calE{{\B{\cal E}}}
\def\bfE{\B{\rm E}}
\def\calG{{\cal G}}
\def\bfG{{\rm G}}
\def\kaplam{{\kappa(\lambda)}}
\def\kaplamp{{\kappa(\lambda')}}
\def\sglasso{{\sc glasso}}
\def\sSMACS{{\sc smacs}}
\def\GL{\sglasso}
\def\bfk{{k}}

\DeclareMathOperator*{\argmin}{arg\,min}

\DeclareMathOperator*{\mini}{minimize}

\def\half{\frac12}

\newcommand{\commentout}[1]{}

\title{\emph{Exact} Covariance Thresholding into Connected Components for large-scale \emph{Graphical Lasso}}
\author{Rahul Mazumder\thanks{email: rahulm@stanford.edu}  \hspace{2cm} Trevor Hastie\thanks{email: hastie@stanford.edu}\\
Department of Statistics\\
Stanford University\\
Stanford, CA  94305.}
\date{Second Draft. Submitted for publication on 8-26-2011 \\
First Draft 8-17-2011.}
\begin{document}
\maketitle

\begin{abstract}%
 We consider the sparse inverse covariance regularization problem or \emph{graphical lasso} with regularization parameter $\lambda$.
Suppose the sample \emph{covariance graph} formed by thresholding the entries of the sample covariance matrix at $\lambda$
is decomposed into connected components. 
We show that the \emph{vertex-partition} induced by the connected components of the thresholded sample covariance graph is \emph{exactly} equal to that induced by the connected components of the estimated concentration graph, obtained by solving the graphical lasso problem.
This characterizes a very interesting property of a path of graphical lasso solutions.
Furthermore, this simple rule, 
when used as a wrapper around existing algorithms for the graphical lasso, leads to enormous performance gains. For a range of values of $\lambda$, our proposal splits a large graphical lasso problem into smaller tractable problems, making it possible to solve an otherwise infeasible large-scale problem. We illustrate the graceful scalability of our proposal via synthetic and real-life microarray examples.
\footnote{The first draft to this paper is available at \url{http://arxiv.org/PS_cache/arxiv/pdf/1108/1108.3829v1.pdf}.}
\end{abstract}

\onehalfspacing

\section{Introduction}\label{intro}
Consider a data matrix $\M{X}_{n \times p}$ comprising of $n$ sample realizations from a $p$ dimensional Gaussian distribution with zero mean and positive definite covariance matrix $\B\Sigma$ (unknown), ie $x_i \stackrel{\mathrm{i.i.d}}{\sim} MVN(\mathbf{0},\B\Sigma)$. The task is to estimate the unknown $\B\Sigma$ based on the $n$ samples.
$\ell_1$ regularized Sparse Inverse Covariance Selection also known as {\emph{graphical lasso}} 
\citep{FHT2007a,BGA2008,yuan_lin_07}
estimates the covariance matrix $\B{\Sigma}$, under the assumption that the inverse covariance matrix i.e.   
$\B{\Sigma^{-1}}$ is sparse. This is achieved by minimizing the regularized negative log-likelihood function:
\begin{equation} \label{eqn-1}
\mini_{\B\Theta \succeq \mathbf{0} } \;\; - \log\det (\B{\Theta} ) + \tr(\s\B{\Theta} ) + \lambda \sum_{i,j}|\B\Theta_{ij}|,
\end{equation} 
where $\M{S}$ is the sample covariance matrix.
Problem (\ref{eqn-1}) is a convex optimization problem in the variable $\B{\Theta}$ \citep{BV2004}. Let 
$\widehat{\B{\Theta}}^{(\lambda)}$ denote the solution to (\ref{eqn-1}). 
We note that (\ref{eqn-1}) can also be used in a more non-parametric 
fashion for any positive semidefinite input matrix $\s$, 
not necessarily a sample covariance matrix of a MVN sample as described above. 

A related criterion to (\ref{eqn-1}) is one where the diagonals are not penalized --- we will study (\ref{eqn-1}) in this paper. 

Developing efficient large-scale algorithms for (\ref{eqn-1}) is an active area of research across the 
fields of Convex Optimization, Machine Learning and Statistics.
Many algorithms have been proposed for this task
\citep[for example]{FHT2007a,BGA2008,Lu:09,Lu:10,Scheinberg_Ma_Goldfarb_2010,sra-11,Yuan_2009,Li:10}.
However, it appears that certain special properties of the solution to (\ref{eqn-1}) have been largely ignored. 
This paper is about one such (surprising) property --- namely establishing an equivalence between the \emph{vertex-partition} 
induced by the connected components of the non-zero patterns of $\widehat{\B{\Theta}}^{(\lambda)}$ and the thresholded
sample covariance matrix $\s$. 
This paper is \emph{not} about a specific algorithm for the problem (\ref{eqn-1}) --- it focuses on the aforementioned 
observation that leads to a novel thresholding/screening procedure based on $\s$. 
This provides interesting insight into the path of solutions $\{\widehat{\B{\Theta}}^{(\lambda)}\}_{\lambda \geq 0}$ obtained by solving (\ref{eqn-1}), over a path of $\lambda$ values. The behavior of the 
connected-components obtained from the non-zero patterns of $\{\widehat{\B{\Theta}}^{(\lambda)}\}_{\lambda \geq 0}$ can be completely understood by simple screening rules on $\s$. This can be done without {\emph{even attempting}}  to 
solve (\ref{eqn-1}) --- arguably a very challenging convex optimization problem.
Furthermore, this thresholding rule can be used as a \emph{wrapper} to enormously boost the performance of existing algorithms, 
as seen in our experiments. 
This strategy becomes extremely effective in solving large problems over a range of values of $\lambda$ --- sufficiently restricted to ensure sparsity and the separation into connected components.

At this point we introduce some notation and terminology, which we will use throughout the paper. 
\subsection{Notations and preliminaries} 
For a matrix $\M{Z}$, its $(i,j)^{\mathrm{th}}$ entry is denoted by $\M{Z}_{ij}$.

We also introduce some graph theory notations and definitions~\citep[see for example]{bollobas-98}.
A finite undirected graph ${\cal G}$ on $p$ vertices is given by the ordered tuple 
${\cal G}=({\cal V}, \B{\cal E})$, where ${\cal V}$ is the set of nodes and $\B{\cal E}$ the collection of (undirected) edges.
The edge-set is equivalently represented via a (symmetric) 0-1 matrix\footnote{0 denotes absence of an edge and 1 denotes its presence.} (also known as the {\emph{adjacency} matrix}) with $p$ rows/columns.
We use the convention that 
a node is not connected to itself, so the diagonals of the adjacency matrix are all zeros.
Let $|{\cal V}|$ and  $|\B{\cal E}|$ denote the number of nodes and edges respectively.

We say two nodes $u,v \in {\cal V}$ are \emph{connected} if there is a \emph{path} between them. A maximal connected \emph{subgraph}\footnote{${\cal G'}=({\cal V'}, \B{\cal E'})$ is a \emph{subgraph} of ${\cal G}$ if ${\cal V'} \subset {\cal V}$ and 
$\B{\cal E'} \subset \B{\cal E}$.} is a \emph{connected component} of the graph ${\cal G}$. \emph{Connectedness} is an equivalence relation that decomposes a graph ${\cal G}$ into 
its connected components $\{({\cal V}_\ell, \B{\cal E}_\ell)\}_{1 \leq \ell \leq K}$ ---
with ${\cal G} = \cup_{\ell=1}^{K} ({\cal V}_\ell, \B{\cal E}_\ell)$, where $K$ denotes the number of connected components.
This decomposition partitions the vertices ${\cal V}$ of ${\cal G}$ into $\{{\cal V}_\ell\}_{1 \leq \ell \leq K}$. 
Note that the labeling of the components is unique upto permutations on $\{1, \ldots, K\}$. 
Throughout this paper we will often refer to this partition as the \emph{vertex-partition} induced by the components of the graph ${\cal G}$. 
If the size of a component is one i.e. $|{\cal V}_\ell|=1$, we say that the node is {\emph{isolated}}.
Suppose a graph $\widehat{\cal G}$ defined on the set of vertices ${\cal V}$ admits the following decomposition into connected components:
$\widehat{\cal G} = \cup_{\ell=1}^{\widehat K} (\widehat{\cal V}_\ell, \widehat{\B{\cal E}}_\ell)$. 
We say the vertex-partitions induced by the  connected components of ${\cal G}$ and $\widehat{\cal G}$ are \emph{equal} if 
$\widehat K =K$ and there is a permutation $\pi$ on $\{1,\ldots, K\}$ such that 
$\widehat{\cal V}_{\pi(\ell)} = {\cal V}_\ell$ for all $\ell \in \{1,\ldots,K\}$. 

The paper is organized as follows. Section \ref{sec:method} describes the covariance graph thresholding idea, along with theoretical justifications and related work, followed by complexity analysis of the algorithmic framework in Section \ref{sec:complexity}.
Numerical experiments appear in Section \ref{sec:numerics}, concluding remarks in Section \ref{sec:con} and the proofs are gathered in the Appendix \ref{sec:appendix}.

\section{Methodology: \emph{Exact} Thresholding of the Covariance Graph}\label{sec:method}
The sparsity pattern of the solution $\widehat{\B{\Theta}}^{(\lambda)}$ to (\ref{eqn-1}) gives rise to the symmetric  edge matrix/skeleton $\in \{0,1\}^{p \times p}$ defined by: 
\begin{equation}\label{conc-graph}
\calE_{ij}^{(\lambda)}= \left\{ \begin{array}{ll}
         1 & \mbox{if $\widehat{\B{\Theta}}^{(\lambda)}_{ij} \neq  0$, $i \neq j$};\\
        0 & \mbox{otherwise}.\end{array} \right. 
\end{equation}
The above defines a symmetric graph  $ \calG^{(\lambda)}  = ({\cal V}, {\calE }^{(\lambda)} )$, namely  
the \emph{estimated concentration graph} \citep{cox-W-96,Laur1996} defined on 
the nodes ${\cal V} = \{1, \ldots, p\}$ with edges ${\calE }^{(\lambda)}$. 

Suppose the graph $\calG^{(\lambda)}$  admits a decomposition into $\kaplam $ connected components:
\begin{equation}\label{decompose-2}
\calG ^{(\lambda)}= \cup_{\ell = 1}^{\kaplam } {\calG }^{(\lambda)}_\ell 
\end{equation} 
where ${\calG }^{(\lambda)}_\ell= (\widehat{\cal V}^{(\lambda)}_\ell,\calE^{(\lambda)}_\ell)$ 
are the components of the graph $\calG ^{(\lambda)}$. 
Note that $\kaplam   \in \{1,\ldots, p\}$, 
with $ \kaplam   = p$ (large $\lambda$) implying that 
all nodes are isolated and for small enough values of $\lambda$, there is only one component i.e.   
$\kaplam  =1$. 

We now describe the simple  screening/thresholding rule. 
Given $\lambda$ we perform a thresholding on the entries of the sample covariance matrix $\s$ and obtain a 
graph edge skeleton ${\bfE }^{(\lambda)} \in \{0,1\}^{p \times p}$ defined by:
\begin{equation}\label{conc-graph-2}
{\bfE }^{(\lambda)}_{ij} = \left\{ \begin{array}{ll}
         1 & \mbox{if $|\s_{ij}| > \lambda$, $i \neq j$};\\
         0 & \mbox{otherwise}.\end{array} \right. 
\end{equation}
The symmetric matrix  ${\bfE }^{(\lambda)}$ defines a symmetric graph on the nodes 
${\cal V} = \{1, \ldots, p\}$ given by $ \bfG ^{(\lambda)} = ({\cal V}, {\bfE }^{(\lambda)})$.
We refer to this as the \emph{thresholded sample covariance graph}. 
Similar to the decomposition in (\ref{decompose-2}), the graph $\bfG ^{(\lambda)}$ also admits a decomposition into connected components:
\begin{equation}\label{decompose-1}
\bfG ^{(\lambda)} = \cup_{\ell = 1}^{\bfk(\lambda)} \bfG ^{(\lambda)}_\ell, 
\end{equation} 
where $\bfG ^{(\lambda)}_\ell = ({\cal V}^{(\lambda)}_\ell,{\bfE }^{(\lambda)}_\ell)$ are the components of the graph $\bfG ^{(\lambda)}$.

Note that the components  of $\calG^{(\lambda)}$ require knowledge of $\widehat{\B{\Theta}}^{(\lambda)}$ --- the solution to (\ref{eqn-1}).
Construction of $\bfG ^{(\lambda)}$ and its components require operating on $\s$ --- an operation that can be performed completely independent of the optimization problem (\ref{eqn-1}), which is arguably more expensive (See Section \ref{sec:complexity}).
The surprising message we describe in this paper is that the \emph{vertex-partition} of the connected components of 
(\ref{decompose-1}) is \emph{exactly} equal to that of (\ref{decompose-2}). 

This observation has the following consequences:
\begin{enumerate}
\item We obtain a very interesting property of the path of solutions $\{\widehat{\B{\Theta}}^{(\lambda)}\}_{\lambda \geq 0}$ ---
the behavior of the connected components of the estimated concentration graph can be completely understood by  
simple screening rules on $\s$ !
\item The cost of computing the connected components of the 
thresholded sample covariance graph (\ref{decompose-1}) is orders of magnitude smaller than the cost of fitting graphical models (\ref{eqn-1}). Furthermore, the computations pertaining to the covariance graph can be done off-line and is amenable to parallel computation (See Section \ref{sec:complexity}).
\item The optimization problem (\ref{eqn-1}) completely separates into $\bfk(\lambda)$ separate optimization sub-problems of the form
(\ref{eqn-1}). The sub-problems have size equal to the number of nodes in each component $p_i := | {\cal V}_i |, i=1,\ldots, \bfk(\lambda)$. 
Hence for certain values of $\lambda$, solving problem (\ref{eqn-1}), 
becomes feasible although it may be impossible to operate on the $p\times p$ dimensional (global) variable $\B{\Theta}$ on a single machine.
\item Suppose that for $\lambda_0$, there are $\bfk(\lambda_0)$ components and the graphical model computations 
are distributed\footnote{Distributing these operations depend upon the number of processors available, their capacities, communication lag, the number of components and the maximal size of the blocks across all machines. These of-course depend upon the computing environment. In the context of the present problem, it is often desirable to club smaller components into a single machine.}. 
 Since the vertex-partitions induced via 
(\ref{decompose-2}) and (\ref{decompose-1}) 
are nested with increasing $\lambda$ (see Theorem \ref{nested-comps-1}), it suffices to operate 
independently on these separate machines to obtain the path of solutions $\{\widehat{\B{\Theta}}^{(\lambda)}\}_{\lambda}$ for all $\lambda \geq \lambda_0$.
\item Consider a distributed computing architecture, where every machine allows operating on a graphical lasso problem (\ref{eqn-1}) of maximal size 
$p_{\max}$. Then with relatively small effort we can find the smallest value of $\lambda=\lambda_{p_{\max}}$, such that there are no connected components of size larger than $p_{\max}$. Problem~(\ref{eqn-1}) thus `splits up'
independently into manageable problems across the different machines. When this structure is not exploited  
the global problem (\ref{eqn-1}) remains intractable.  
\end{enumerate}
The following theorem establishes the main technical contribution of this paper---the equivalence of the 
vertex-partitions induced by the connected components of the thresholded sample covariance graph and the estimated concentration graph.
\begin{thm}\label{thm-equal-1}
For any $\lambda >0$, the components of the estimated concentration graph  
$\calG ^{(\lambda)}$, as defined in (\ref{conc-graph}) and (\ref{decompose-2}) induce \emph{exactly} the same vertex-partition 
as that of the thresholded sample covariance graph $\bfG ^{(\lambda)}$, defined in (\ref{conc-graph-2}) and (\ref{decompose-1}). 
That is $\kaplam =\bfk(\lambda)$ and there exists a permutation
$\pi$ on $\{1,\ldots, \bfk(\lambda)\}$ such that: 
\begin{equation}\label{eq-conn-comp}
\widehat{\cal V}^{(\lambda)}_i={\cal V}^{(\lambda)}_{\pi(i)}, \;\; \forall i =1, \ldots, \bfk(\lambda).
\end{equation}
\begin{proof}
The proof of the theorem appears in Appendix~\ref{proof-thm1}.
\end{proof}
\end{thm}
Since the decomposition of a symmetric graph 
into its connected components depends upon the ordering/ labeling  of the components, 
the permutation $\pi$ appears in Theorem \ref{thm-equal-1}.    

\begin{rem}
Note that the edge-structures \emph{within} each block need not be preserved. Under a matching reordering of the labels 
of the components of $\calG ^{(\lambda)}$ and $\bfG ^{(\lambda)}$:\\
for every 
fixed $\ell$ such that $\widehat{\cal V}^{(\lambda)}_{\ell}={\cal V}^{(\lambda)}_{\ell}$ the edge-sets
 ${\calE}^{(\lambda)}_{\ell}$  and ${\bfE }^{(\lambda)}_{\ell}$ are \emph{not} 
necessarily equal.
\end{rem}

Theorem~\ref{thm-equal-1} leads to a special property of the path-of-solutions to (\ref{eqn-1}), i.e.   the 
vertex-partition induced by the connected components of $\calG ^{(\lambda)}$ are nested with increasing $\lambda$. This is the 
content of the following theorem.

\begin{thm}\label{nested-comps-1}
Consider two values of the regularization parameter such that 
$\lambda > \lambda' >0$, with corresponding concentration graphs $\calG ^{(\lambda)}$ and $\calG ^{(\lambda')}$ as in (\ref{conc-graph}) and connected components (\ref{decompose-2}).
Then the vertex-partition induced by the components of $\calG ^{(\lambda)}$ 
are \emph{nested} within the partition induced by the components of $\calG ^{(\lambda')}$. 
Formally, $\kaplam  \geq \kaplamp $ and the vertex-partition 
$\{{\widehat{\cal V}}^{(\lambda)}_{\ell}\}_{1 \leq \ell \leq \kaplam }$ forms a finer resolution of $\{{\widehat{\cal V}}^{(\lambda')}_{\ell}\}_{1 \leq \ell \leq \kaplamp }$. 
\begin{proof}
The proof of this theorem appears in the Appendix~\ref{proof-thm2}.
\end{proof} 
\end{thm}
\begin{rem}\label{rem-2}
It is worth noting that Theorem \ref{nested-comps-1} addresses the nesting of the edges { \emph{across} } connected components and not within a component.
In general, the edge-set ${\calE}^{(\lambda)}$ of the estimated concentration graph need not be nested as a function of $\lambda$:\\
for $\lambda > \lambda'$, in general, ${\calE}^{(\lambda)} \not \subset {\calE}^{(\lambda')}$. 
\end{rem}
See \citet[Figure 3]{FHT2007a}, for numerical examples demonstrating the non-monotonicity of the edge-set across $\lambda$, as described in
Remark \ref{rem-2}.

\subsection{Related Work}\label{sec:related-work}
\citet{witten-11} fairly recently proposed a scheme to detect \emph{isolated} 
nodes for problem (\ref{eqn-1}) via a simple screening of the entries of $\s$. Using the notation in \citet[Algorithm 1]{witten-11}, 
the authors propose operating criterion (\ref{eqn-1}) on the set of non-isolated nodes (obtained from the sample covariance matrix) i.e.
$\{1,\ldots,p\} \setminus {\cal C}$, where the isolated nodes are given by ${\cal C}$: 
\begin{equation}\label{node-screen}
{\cal C} = \{ i : |\s_{ij}| \leq \lambda, \forall j \neq i\}. 
\end{equation}
The authors showed that ${\cal C}$ is exactly equivalent to the set of isolated nodes 
of the estimated precision matrix obtained by solving (\ref{eqn-1}) on the entire $p\times p$ dimensional problem.
Earlier, \citet{BGA2008}[Theorem 4] also made the same observation.
This is of-course \emph{related} to a very special case of the proposal in this paper. 
Suppose in Theorem~\ref{thm-equal-1}, the estimated concentration graph admits a decomposition where 
some of the connected components have size one --- \citet{witten-11} only screens  
the isolated nodes and treats the the remaining nodes as a separate `connected unit'. Although the original version of their paper 
\citep{witten-11} deal with single nodes, we have learned \citep{witten-update} that with N. Simon they have also 
discovered a form of  block screening. 

This node-screening strategy (\ref{node-screen}) was used by them  as a wrapper around the (graphical lasso) {\GL} algorithm of \citet{FHT2007a} --- leading to 
substantial improvements over the existing {\GL} solver of \citet{FHT2007a} (CRAN {\GL}  package version 1.4). 
However, we show below that  
the node-screening idea is actually an immediate \emph{consequence} of the block coordinate-wise updates used by the {\GL} algorithm --- 
an observation that was not exploited by the solver.

Recall that the {\GL} algorithm \citep{FHT2007a} operates in a block-coordinate-wise i.e. row/column fashion on 
the variable $\M{W}=\B{\Theta}^{-1}$. 
The method partitions the problem variables as follows: 
\begin{eqnarray}\label{break-x}
\B{\Theta} = \left(
  \begin{array}{cc}
    \B{\Theta}_{11} & \B{\theta}_{12} \\
    \B{\theta}_{21}  & \theta_{22} \\
  \end{array}
\right),                 & \M{S} = \left( 
  \begin{array}{cc}
    \mathbf{S}_{11} & \M{s}_{12} \\
    \M{s}_{21}  & s_{22} \\
  \end{array}
\right), & \M{W} = \left(
\begin{array}{cc}
    \M{W}_{11} & \M{w}_{12} \\
    \M{w}_{21}  & w_{22} \\
  \end{array}
\right)
\end{eqnarray}
where  the last row/ column represents the optimization variable, the others being fixed. 
The partial optimization problem w.r.t. the last row/column (leaving apart the diagonal entry) is given by:
\begin{equation} \label{lasso-grad-3}
\widehat{\B{\theta}}_{12}:= \argmin_{\B{\theta}_{12}}\;\;\left\{ \half \B{\theta}_{12}'\M{W}_{11}\B{\theta}_{12} +  \B{\theta}'_{12}\theta_{22} \M{s}_{12} + \lambda \theta_{22}\|\B{\theta}_{12}\|_1 \right\}.
\end{equation}
Clearly the solution $\widehat{\B{\theta}}_{12}$ of the above (\ref{lasso-grad-3}) is zero iff
\begin{equation}\label{partial-screen-1}
\|\M{s}_{12}\|_\infty \leq \lambda
\end{equation}
--- a condition depending \emph{only} on that row/column of $\s$.
As we pointed out before, the above condition (\ref{partial-screen-1}) is \emph{exactly} the condition for node-screening 
(\ref{node-screen}) described in \citet{witten-11}. 
The notable improvement in timings observed in \citet{witten-11} with node screening goes on to suggest that 
the {\GL} solver of \citet{FHT2007a} (as implemented in CRAN { \sglasso\ }  package Version 1.4)
does \emph{not} make the check (\ref{partial-screen-1}), before going on to solve
problem (\ref{lasso-grad-3}). The existing implementation goes on to optimize (\ref{lasso-grad-3}) --- a
$\ell_1$ regularized quadratic program via cyclical coordinate-descent. Note that (\ref{lasso-grad-3}), in its own right, 
is fairly challenging to solve for large problems. 
\section{Computational Complexity}\label{sec:complexity}
The overall complexity of our proposal depends upon (a) the graph partition stage and (b) solving (sub)problems of the form 
(\ref{eqn-1}). In addition to these, there is an unavoidable complexity associated with handling and/or forming $\s$.

The cost of computing the connected components of the 
thresholded covariance graph is fairly negligible when compared to solving a similar sized graphical lasso problem (\ref{eqn-1}) --- see also our simulation studies in Section \ref{sec:numerics}. 
In case we observe samples $x_i \in \Re^{p},i =1,\ldots,n$ the cost for creating the sample covariance matrix $\s$ is $O(n\cdot p^2)$.
 Thresholding the sample covariance matrix  costs  $O(p^2)$.
Obtaining the connected components of the thresholded covariance graph costs 
$O( |{\bfE }^{(\lambda)}| + p)$ \citep{tarjan-72}. Since we are interested in a region where the thresholded covariance graph is sparse enough to be broken into smaller connected components --- $|{\bfE }^{(\lambda)}| \ll p^2$. 
Note that all computations pertaining to the construction of the connected 
components and the task of computing $\s$ can be computed off-line. Furthermore the computations are parallelizable.  
\citet[for example]{par-conn-comp-91} describes parallel algorithms for computing connected components of a graph --- they 
have a time complexity $O(\log p)$ and require $O((|{\bfE }^{(\lambda)}|+ p)/\log(p))$ processors with space 
$O(p+ |{\bfE }^{(\lambda)}|)$.

There are a wide variety of algorithms for the task of solving (\ref{eqn-1}). While an exhaustive review of the computational complexities of the different algorithms is beyond the scope of this paper, we provide a brief summary for a few algorithms below.

\citet{BGA2008} proposed a smooth accelerated gradient based method \citep{nest_05} with complexity $O(\frac{p^{4.5}}{\epsilon})$ to obtain an $\epsilon$ accurate solution --- the per iteration cost being $O(p^3)$. They also proposed a block coordinate method which has a complexity of $O(p^4)$.

The complexity of the {\GL} algorithm \citep{FHT2007a} which uses a row-by-row block coordinate method is roughly $O(p^3)$ for reasonably sparse-problems with $p$ nodes. For denser problems the cost can be as large as $O(p^4)$.

The algorithm \textsc{smacs} proposed in \citet{Lu:10} has a 
per iteration complexity of $O(p^3)$ and an overall complexity of $O(\frac{p^4}{\sqrt{\epsilon}})$ to obtain an
$\epsilon > 0$ accurate solution.

It appears that most existing algorithms for (\ref{eqn-1}), have a complexity of at least $O(p^3)$ to $O(p^4)$ or possibly larger, depending upon the algorithm used and the desired accuracy of the solution --- making computations for (\ref{eqn-1}) almost impractical for values of $p$ much larger than 2000.

It is quite clear that the role played by covariance thresholding is indeed crucial in this context. 
Assume that we choose to use a solver of complexity $O(p^J)$, with $J \in \{3,4\}$, along with 
our screening procedure.
Suppose for a given $\lambda$ the thresholded sample covariance graph has $\bfk(\lambda)$ components --- the total cost of solving these smaller problems is then $\sum_{i=1}^{\bfk(\lambda)} O(| {\cal V}^{(\lambda)}_i |^J) \ll O(p^J)$, with $J \in \{3,4\}$. 
This difference in practice can be enormous --- see Section \ref{sec:numerics} for numerical examples. 
This is what makes large scale graphical lasso problems solvable !

\section{Numerical examples}\label{sec:numerics}
In this section we show via numerical experiments that the screening property helps in obtaining many fold speed-ups when compared to 
an algorithm that does not exploit it. Section \ref{sec:synthetic} considers synthetic examples and  
Section \ref{sec:real} discusses real-life microarray data-examples.

\subsection{Synthetic examples}\label{sec:synthetic}
Experiments are performed with two publicly available algorithm implementations for the problem (\ref{eqn-1}): 
\begin{description}
\item [\sglasso:] The algorithm of \citet{FHT2007a}. We used the MATLAB wrapper available at 
\url{http://www-stat.stanford.edu/~tibs/glasso/index.html} to the Fortran code. 
The specific criterion for convergence (lack of progress of the diagonal entries) was set to $10^{-5}$ and the maximal number of iterations was set to 1000.
\item[\sSMACS:] denotes the algorithm of \citet{Lu:10}. We used the  MATLAB  implementation \verb$smooth_covsel$ 
available at \url{http://people.math.sfu.ca/~zhaosong/Codes/SMOOTH_COVSEL/}.
The criterion for convergence (based on duality gap) was set to $10^{-5}$ and the maximal number of iterations was set to 1000.
\end{description}
We will like to note that the convergence criteria of the two algorithms \sglasso\ and \sSMACS\ are not the same.
For obtaining the connected components of a symmetric adjacency matrix we used the MATLAB function \verb$graphconncomp$.
All of our computations are done in MATLAB 7.11.0 on a 3.3 GhZ Intel Xeon processor.

The simulation examples are created as follows. We generated a block diagonal 
matrix given by
$\tilde{\s}=\mathrm{blkdiag}(\tilde{\s}_1,\ldots, \tilde{\s}_K)$, where each block
$\tilde{\s}_\ell = \M{1}_{p_{\ell} \times p_{\ell}}$ --- a matrix of all ones and $\sum_\ell p_\ell =p$. 
In the examples we took all $p_\ell$'s to be equal to $p_1$ (say).  
Noise of the form 
$\sigma \cdot UU'$ ($U$ is a $p \times p$ matrix with i.i.d. standard Gaussian entries) 
is added to $\tilde{\s}$ such that  1.25 times the largest (in absolute value) off block-diagonal (as in the block structure of $\tilde{\s}$)
entry of $\sigma\cdot UU'$
equals the smallest absolute non-zero entry in $\tilde{\s}$ i.e. one.
The sample covariance
matrix is  $\s = \tilde{\s} +  \sigma \cdot UU'$. 

We consider a number of examples for varying $K$ and $p_1$ values, as shown in Table \ref{table-1}. Sizes were chosen such that 
it is at-least `conceivable' to solve (\ref{eqn-1}) on the full dimensional problem, without screening.
In all the examples shown in Table \ref{table-1}, we set $\lambda_{I}:=(\lambda_{\max} + \lambda_{\min})/2$, where for all values of $\lambda$ in the interval $[\lambda_{\min},\lambda_{\max}]$ the thresh-holded version of the sample covariance matrix has exactly $K$ connected components. We also took a larger value of $\lambda$ i.e.   $\lambda_{II}:=\lambda_{\max}$, which gave sparser estimates of the precision matrix 
but the number of connected components were the same.
 
The computations across different connected blocks could be distributed into as many machines. 
This would lead to almost a $K$ fold improvement in timings, however in Table \ref{table-1} we report the timings by operating 
serially across the blocks. The serial `loop' across the different blocks are implemented in MATLAB.

\begin{table}
\begin{tabular}{ l c c c c  c c c }\hline
    \multirow{3}{*}{K} &  \multirow{3}{*}{$p_1$ / p }       &      \multirow{3}{*}{$\lambda$}          & \multirow{3}{*}{\textbf{Algorithm}}   &  \multicolumn{2}{c}{\textbf{Algorithm Timings} (sec)} & \textbf{Ratio}& \textbf{Time} (sec) \\
 &  &       &      & with& without & Speedup & graph \\
 &  &       &      & screen &  screen & factor & partition \\ \hline \hline
2 &      200  / 400   &  \multirow{2}{*}{$\lambda_I$} & { \sglasso }    &    11.1  & 25.97 & 2.33 &  \multirow{2}{*}{0.04} \\
    &                    &   &  { \sSMACS }   & 12.31 & 137.45 & 11.16 & \\ \\
&          &  \multirow{2}{*}{$\lambda_{II}$} &  \sglasso     &   1.687 &  4.783 & 2.83 & \multirow{2}{*}{0.066}\\
   &                    &   & { \sSMACS }   & 10.01 &  42.08 & 4.20 & \\ \hline
2  &   500  /1000  & \multirow{2}{*}{$\lambda_I$} & { \sglasso }     &   305.24 &  735.39  & 2.40 & \multirow{2}{*}{0.247}\\
    &                &   &  { \sSMACS }   &   175  &  2138*  & 12.21 &  \\ \\
  &     &  \multirow{2}{*}{$\lambda_{II}$} & { \sglasso }     &    29.8  &  121.8 & 4.08 &\multirow{2}{*}{0.35}\\
    &     &      &  { \sSMACS }   &   272.6 &   1247.1  &   4.57 &\\ \hline
5  &   300  /1500  &  \multirow{2}{*}{$\lambda_I$} & { \sglasso }    &  210.86 &  1439  & 6.82 & \multirow{2}{*}{0.18} \\
    &                   &  &  { \sSMACS }   &   63.22 &  6062*  & 95.88 &\\ \\
  &   &  \multirow{2}{*}{$\lambda_{II}$}& { \sglasso }    &  10.47& 293.63 & 28.04 &  \multirow{2}{*}{0.123}\\
    &                   &  &  { \sSMACS }   &   219.72&  6061.6 &  27.58 &\\ \hline
5  & 500 /2500      & \multirow{2}{*}{$\lambda_I$} &  { \sglasso }    & 1386.9 & -  & - &\multirow{2}{*}{0.71} \\
   &                 &  & { \sSMACS }   &  493 &  -  & - & \\ \\
 &           &  \multirow{2}{*}{$\lambda_{II}$} &  { \sglasso }    & 17.79 & 963.92   & 54.18 & \multirow{2}{*}{0.018} \\
   &                 &  & { \sSMACS }   &  354.81 & - & - &\\ \hline
8 &   300  /2400  &  \multirow{2}{*}{$\lambda_I$} &  { \sglasso }    & 692.25 &   -  & - &  \multirow{2}{*}{0.713} \\ 
    &                 &   &  { \sSMACS }   &   185.75 &  -  & - &\\ \\
&     &  \multirow{2}{*}{$\lambda_{II}$}&  { \sglasso }    &9.07 &   842.7  & 92.91 &\multirow{2}{*}{0.023}\\
    &                 &   &  { \sSMACS }   & 153.55 & - & - &\\ \hline
\end{tabular}
\caption{Table showing (a) the times in seconds with screening, (b)  without screening i.e. on the whole matrix and (c) 
the ratio (b)/(a) -- `Speedup factor' for algorithms {\sglasso} and { \sSMACS }. 
Algorithms with screening are operated serially---the times reflect the total time summed across all blocks.
The column `graph partition' lists the time for computing the connected components of the thresholded sample covariance graph. 
Since $\lambda_{II} > \lambda_{I}$, the former gives sparser models. `*' denotes the algorithm did not converge within 1000 iterations. `-' refers to cases where the respective algorithms failed to converge within 2 hours.}  \label{table-1}
\end{table}

Table \ref{table-1} shows the rather remarkable improvements  obtained by using our proposed covariance thresholding strategy as compared to operating on the whole matrix.
Timing comparisons between \sglasso\  and { \sSMACS } are not fair, since \sglasso\  is written in Fortran and  { \sSMACS } in MATLAB. However, we note that our experiments are meant to demonstrate how the thresholding helps in improving the overall computational time over the baseline method of not exploiting screening.  
It is interesting to observe that there is almost a role-reversal in the performances of \sglasso\  and { \sSMACS } with changing $\lambda$ values, for the cases with screening. 
$\lambda_I$ corresponds to a denser solution of the precision matrix --- here \sglasso\  converges more slowly than { \sSMACS }. For larger values of the tuning parameter i.e.   $\lambda = \lambda_{II}$, the solutions are sparser --- \sglasso\  converges much faster 
than {\sSMACS}.
For problems without screening we observe that \sglasso\ converges much faster than \textsc{smacs}, for both values of the tuning parameter.
This is probably because of the intensive matrix computations associated with the \textsc{smacs} algorithm.   
 Clearly our proposed strategy makes solving larger problems 
(\ref{eqn-1}), not only feasible but with quite attractive computational time. The time taken by 
the graph-partitioning step in splitting the thresholded covariance graph into its connected components is negligible as compared to the timings for the optimization problem.

\subsection{Micro-array Data Examples}\label{sec:real}
The graphical lasso is often used in learning connectivity networks in gene-microarray data \cite[see for example]{FHT2007a}.
Since in most real examples the number of genes $p$ is around tens of thousands, obtaining an inverse covariance matrix by solving (\ref{eqn-1}) is computationally impractical. The covariance thresholding method we propose easily applies to these problems --- and as we see gracefully delivers solutions over a large range of the parameter $\lambda$. 
We study three different micro-array examples and observe that as one varies $\lambda$ from large to small values, 
the thresholded covariance graph splits into a number of non-trivial connected components of varying sizes. 
We continue till a small/moderate  value of $\lambda$ when the maximal size of a connected component gets larger than a predefined 
machine-capacity or the `computational budget' for a single graphical lasso problem. 
Note that in relevant micro-array applications, since $p\gg n$ ($n$, the number of samples is at most a few hundred) 
heavy regularization is required to control the variance of the covariance estimates --- so it does seem 
reasonable to restrict to solutions of (\ref{eqn-1}) for large values of $\lambda$.

Following are the data-sets we used for our experiments:
\begin{enumerate}
\item[(A)] This data-set appears in \citet{AB1999} and has been analyzed by
\citet[for example]{rothman_2010}.
In this experiment, tissue samples were analyzed using an Affymetrix oligonucleotide array. The data were processed, filtered and
reduced to a subset of $p=2000$ gene expression values. The number of colon adenocarcinoma tissue samples is $n=62$.
\item[(B)] This is an early example of an expression array, obtained
from the Patrick Brown lab at Stanford University. There are $n=385$
patient samples of tissue from various regions of the body (some from tumors, some not), with
gene-expression measurements for $p=4718$ genes. 
\item[(C)] The third example is the by now famous NKI dataset that produced
the 70-gene prognostic signature for breast cancer \citet{pmid12490681}. Here there
are $n=295$ samples and  $p=24481$ genes.
\end{enumerate}
Among the above, both (B) and (C) have few missing values --- which we imputed by the respective global means of the 
observed expression values. For each of the three data-sets, we took $\s$ to be the corresponding sample correlation matrix. 

\begin{figure}[ht]
  \centering
 \begin{psfrags}
 \psfrag{ALONDATA}[][ct]{\textsc{(A)} \hspace{.01in} \scriptsize{p=2000} }
 \psfrag{RHO}[][t]{\large{$\lambda$}}
\psfrag{ConnCompSizes}[][cb]{\small{$\log_{10}$(Size of Components)}}
\includegraphics[width=.31\textwidth,height=3in]{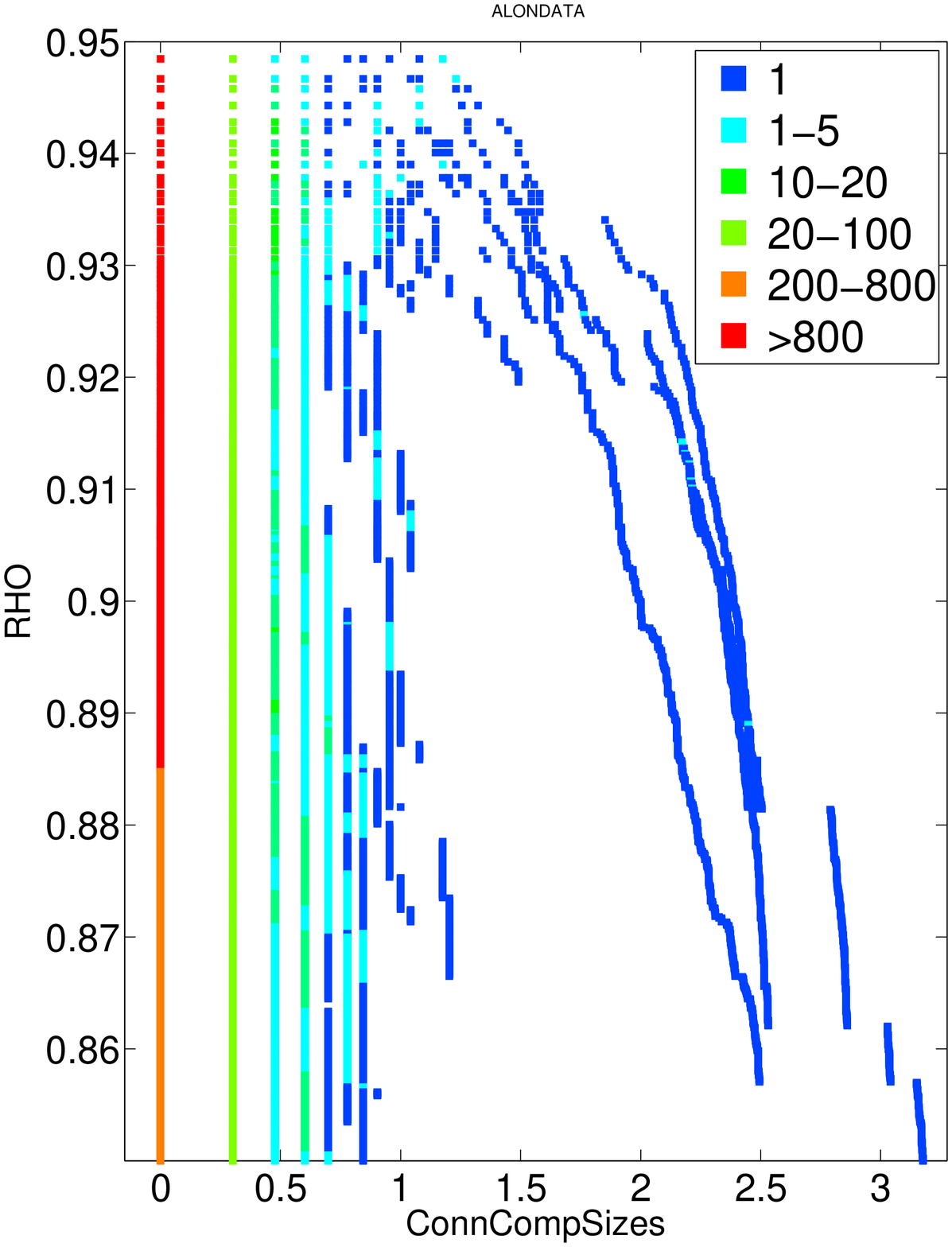}
\end{psfrags}
\begin{psfrags}
 \psfrag{BROWNDATA}[][ct]{\textsc{(B)}  \hspace{.01in} \scriptsize{p=4718} }
 \psfrag{RHO}{}
\psfrag{ConnCompSizes}[][cb]{\small{$\log_{10}$(Size of Components)}}
 \includegraphics[width=.31\textwidth,height=3in]{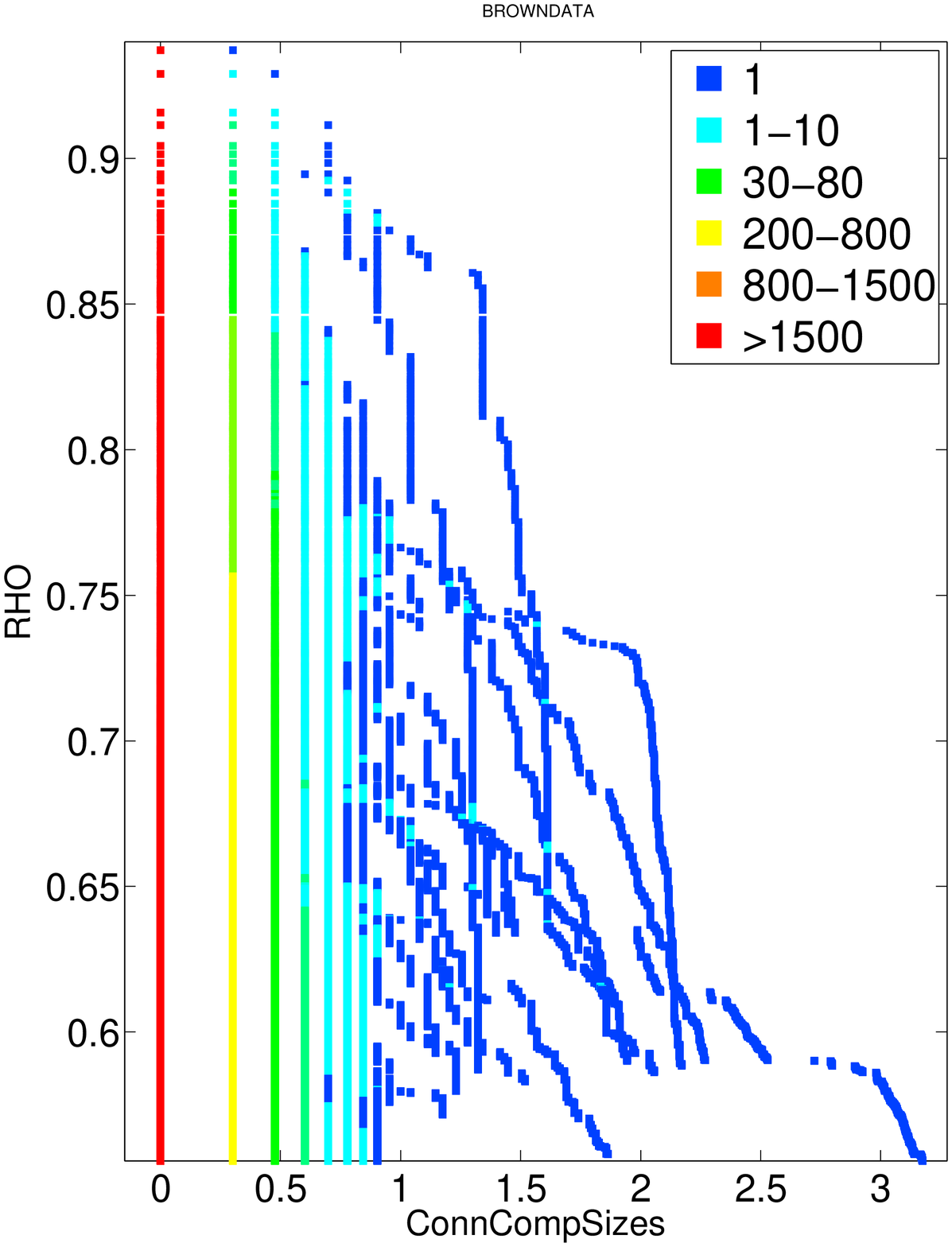}
\end{psfrags}
\begin{psfrags}
 \psfrag{NKIDATA}[][ct]{\textsc{(C)}  \hspace{.01in} \scriptsize{p=24281} }
 \psfrag{RHO}{}
\psfrag{ConnCompSizes}[][cb]{\small{$\log_{10}$(Size of Components)}}
 \includegraphics[width=.31\textwidth,height=3in]{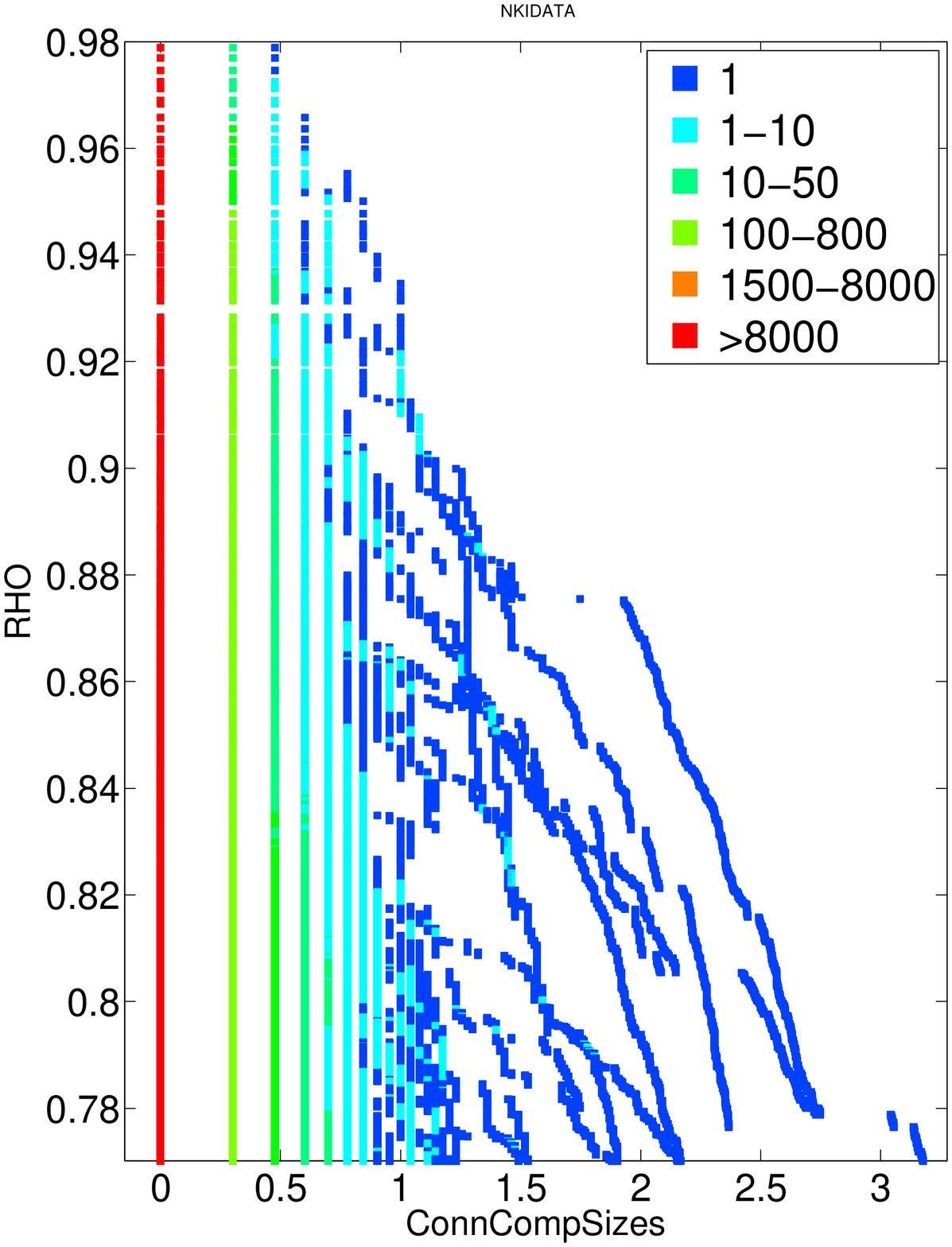}
 \end{psfrags}
  \caption{Figure showing the size distribution (in the log-scale) of connected components arising from the thresholded sample covariance graph for examples (A)-(C). For every value of $\lambda$ (vertical axis), the horizontal slice denotes the sizes of the different components appearing in the thresholded covariance graph. The colors represent the number of components in the graph having that specific size. 
For every figure, the range of $\lambda$ values is chosen such that the maximal size of the connected components do not exceed 1500.}
\label{fig-conn-comp}
\end{figure}

Figure \ref{fig-conn-comp} shows how the component sizes of the thresholded covariance graph change across $\lambda$.
We describe the strategy we used to arrive at the figure. 
Note that the connected components change {\emph{only}} at the absolute values of the entries of $\s$.
From the sorted absolute values of the off-diagonal entries of $\s$, we 
obtained the smallest value of $\lambda$, say $\lambda'_{\min}$, 
for which the size of the maximal connected component was 1500. For a grid of values of 
$\lambda$ till $\lambda'_{\min}$, we computed the connected components 
of the thresholded sample-covariance matrix and obtained the size-distribution of the various connected components. 
Figure \ref{fig-conn-comp} shows how these components change over a range of values of $\lambda$ for the three examples (A), (B) and (C).
The number of connected components of a particular size is denoted by a color-scheme, described by the color-bar 
in the figures. With increasing $\lambda$: the larger connected components gradually disappear as they decompose into smaller components; 
the sizes of the connected components decrease and the frequency of the smaller components increase. Since these are all correlation matrices, for $\lambda\geq 1$ all the nodes in the graph become isolated. The range of $\lambda$ values for which the maximal size of the components is smaller than 1500 differ across the three examples. For (C) there is a greater variety in the sizes of the components as compared to (A) and (B). 
Note that by Theorem \ref{thm-equal-1}, the pattern of the components appearing in Figure \ref{fig-conn-comp} are exactly the same as the components appearing in the solution of (\ref{eqn-1}) for that $\lambda$.

Continuing from Table \ref{table-1}, we proceed to show that the screening rules lead to encouraging speed-ups for real-data examples as well. 
We consider example (A) and apply on it \sglasso\  and {\sSMACS} with and without screening on a grid of $\lambda$ values.
For smaller values of $\lambda$ in the range, 
\sglasso\ and {\sSMACS} take a very long time to converge. 
Comparative timings appear in Table \ref{table-2}.
In these experiments we took the criterion of convergence for both  { \sglasso\ } and {\sSMACS}   
as $10^{-4}$ and they were run till a maximum of 500 iterations. 
The algorithms were run independently across the grid of $\lambda$ values chosen.
The times displayed for the algorithms with screening indicate the total time required by solving the smaller sub-problems serially
--- the results shown are to emphasize the speed-ups obtained in each of the algorithms via screening.

\begin{table}
\begin{tabular}{c c  c c c c}\hline
 \textbf{\small{Average size} } & \multirow{3}{*}{\textbf{Algorithm}}   &  \multicolumn{2}{c}{\textbf{Algorithm Timings} (sec)} & \textbf{Ratio} & \textbf{Time} (sec) \\
 \textbf{\small{of maximal}}& & with& without & Speedup & graph \\
 \textbf{ \small{component} }&  & screen &  screen & factor & partition \\ \hline \hline
   \multirow{2}{*}{5} & { \sglasso }    &  0.02   & 3866 & $1.9\times 10^5$ &  \multirow{2}{*}{0.009} \\ 
    &  { \sSMACS }   &  0.87 & $1.16\times 10^5$  &   $1.33\times 10^5$ & \\ \\
 \multirow{2}{*}{727} & { \sglasso }    & 413 & 13214  &   32  &  \multirow{2}{*}{0.14} \\
 &  { \sSMACS }   & 4285 & $2.7\times 10^5$ &   63 & \\ \hline
\end{tabular}
\caption{Timings for Eg (A): table showing times with/without screening,  
`Speedup factor' and time for `graph partition' as in Table \ref{table-1}, for two different ranges of $\lambda$-values.
Here $p=2000$ and the times for each of the two columns are summed over 10 different $\lambda$ values. 
The left-most column is the size of the maximal connected component, averaged across the $\lambda$ values. 
We see that the times increase with decreasing sparsity and the speed-up factor is impressively large when there are a large number of small-sized connected components. The cost of computing the components of the thresholded covariance graph is relatively negligible.}  \label{table-2}
\end{table}
For examples (B) and (C) the full problem sizes are beyond the scope of { \sglasso\ } and {\sSMACS} --- 
the screening rule is apparently the \emph{only} way to obtain solutions for a reasonable range of $\lambda$-values as shown in 
Figure \ref{fig-conn-comp}. We report in Table \ref{table-3} the averaged time
taken by each of  { \sglasso\ } and {\sSMACS} over a grid of 100 $\lambda$-values, for examples (B) and (C).
The 100 $\lambda$ values correspond to the top 2 \% sorted absolute values of the off-diagonal entries in $\s$ below $\lambda_{500}$ --- where 
$\lambda_{500}$ is the smallest value of $\lambda$ such that the maximal component in the thresholded covariance graph has size 500.
\begin{table}
\begin{tabular}{c c  c c c}\hline
\multirow{2}{*}{\textbf{Example} / $p$} & { \textbf{\small{Average size of}} } & \multicolumn{2}{c}{\textbf{Algorithm Timings} (sec)} & \textbf{Time}(sec) \\
     & {\textbf{\small{maximal component}} }& { \sglasso } & { \sSMACS }  & graph-partition \\ \hline\hline
(B) / 4718    & 330 & 4.93 & 31.09 & 0.0082 \\  \\
(C) / 24481   & 461 & 15.4 & 141 & 0.0186 \\ \hline
\end{tabular}
\caption{Averaged timings (in secs) for different algorithms for examples (B) and (C) over a grid of 100 $\lambda$ values, as described in the text. Algorithms are applied with the screening rule.}  \label{table-3}
\end{table}

\section{Conclusions}\label{sec:con}
In this paper we present a novel property characterizing the family of solutions to the graphical lasso problem (\ref{eqn-1}), as a function of the regularization parameter $\lambda$. The property is fairly surprising --- the 
vertex partition induced by the connected components of the non-zero patterns of the estimated concentration matrix and the thresholded
sample covariance matrix $\s$ are \emph{exactly equal}. 
This property seems to have been unobserved in the literature.
Our observation not only provides interesting insights into
the properties of the graphical lasso solution-path but also opens the door to solving large-scale graphical lasso problems, 
which are otherwise intractable. 
This simple rule when used as a wrapper around existing algorithms leads to enormous performance boosts --- on occasions by a factor of thousands!

\appendix
\section{Proofs}\label{sec:appendix}
\subsection{Proof of Theorem \ref{thm-equal-1}} \label{proof-thm1}
\begin{proof}
Suppose $\widehat{\B{\Theta}}$ (we suppress the superscript $\lambda$ for notational convenience)
solves problem (\ref{eqn-1}), then standard KKT conditions of optimality \citep{BV2004} give: 
\begin{eqnarray}
| \s_{ij}-\widehat{\M{W}}_{ij}| \leq \lambda\;& \forall \;\widehat{\B{\Theta}}_{ij} =0;&\;\;\; \text{and}\label{kkt-1}\\
\widehat{\M{W}}_{ij} =  \s_{ij} + \lambda\; &\forall\; \widehat{\B{\Theta}}_{ij} > 0; & \widehat{\M{W}}_{ij} =  \s_{ij} - \lambda\; 
\forall \; \widehat{\B{\Theta}}_{ij} <0; \label{kkt-2}
\end{eqnarray}
where $\widehat{\M{W}} = (\widehat{\B{\Theta}})^{-1}$. The diagonal entries satisfy 
$\widehat{\M{W}}_{ii} = \s_{ii} + \lambda$, for  $i =1,\ldots,p$. 

Using (\ref{conc-graph-2}) and (\ref{decompose-1}), there exists an ordering of the vertices $\{1,\ldots, p\}$ of the graph 
 such that $\bfE ^{(\lambda)}$ is block-diagonal. 
For notational convenience, we will assume that the matrix is already in that order. 
Under this ordering of the vertices, the edge-matrix of the thresholded covariance graph is of the form: 
\begin{equation}\label{matrix-block-diag1}
{\bfE }^{(\lambda)} = \begin{pmatrix} {\bfE }_{1}^{(\lambda)}& 0 & \cdots &  0 \\ 
0 & {\bfE }_{2}^{(\lambda)} & 0 & \cdots  \\ 
\vdots & \vdots & \ddots & \vdots \\ 
0 & \cdots & 0& {\bfE }_{\bfk(\lambda)}^{(\lambda)}  \end{pmatrix}
\end{equation}
where the different components represent blocks of indices given by:
${\cal V}^{(\lambda)}_\ell, \ell=1,\ldots, \bfk(\lambda)$.

We will construct a matrix $\widehat{\M{W}}$  having the same structure as (\ref{matrix-block-diag1}) which 
is a solution to (\ref{eqn-1}). Note that if  $\widehat{\M{W}}$ is block diagonal then so is its inverse. 
Let $\widehat{\M{W}}$ and its inverse $\widehat{\B{\Theta}}$ be given by:
\begin{equation}\label{matrix-block-diag2}
\widehat{\M{W}} = \begin{pmatrix} \widehat{\M{W}}_{1} & 0 & \cdots &  0 \\ 
0 & \widehat{\M{W}}_{2} & 0 & \cdots  \\ 
\vdots & \vdots & \ddots & \vdots \\ 
0 & \cdots & 0& \widehat{\M{W}}_{\bfk(\lambda)} \end{pmatrix},\;\; \;\;\widehat{\B{\Theta}} 
= \begin{pmatrix} \widehat{\B{\Theta}}_{1} & 0 & \cdots &  0 \\ 
0 &  \widehat{\B{\Theta}}_{2} & 0 & \cdots  \\ 
\vdots & \vdots & \ddots & \vdots \\ 
0 & \cdots & 0&  \widehat{\B{\Theta}}_{\bfk(\lambda)} \end{pmatrix}
\end{equation}
Define the block diagonal matrices $\widehat{\M{W}}_{\ell}$ or equivalently $\widehat{\B{\Theta}}_{\ell}$ via 
the following sub-problems
\begin{equation} \label{sub-block-gl}
\widehat{\B{\Theta}}_{\ell} = \argmin_{\B{\Theta}_\ell} \;\;\;\{ -\log\det(\B{\Theta}_\ell) + 
\tr(\s_\ell \B{\Theta}_\ell) + \lambda \sum_{ij}|\left(\B{\Theta}_{\ell}\right)_{ij}| \}
\end{equation}
for $\ell = 1,\ldots,\bfk(\lambda)$, where
$\s_\ell$ is a sub-block of $\s$, with row/column 
indices from ${\cal V}_\ell^{(\lambda)} \times {\cal V}_\ell^{(\lambda)}$. The same notation is used for $\B\Theta_\ell$.
Denote the inverses of the block-precision matrices by $\{\widehat{\B{\Theta}}_{\ell}\}^{-1} = \widehat{\M{W}}_\ell$.
We will show that the above $\widehat{\B{\Theta}}$ satisfies the KKT conditions ---~(\ref{kkt-1}) and (\ref{kkt-2}). 

Note that by construction of the thresholded sample covariance graph, \\
if $i \in {\cal V}^{(\lambda)}_\ell$ and $j \in  {\cal V}^{(\lambda)}_{\ell'}$ with $\ell \neq \ell'$, then 
$|\s_{ij}| \leq \lambda$. 

Hence, for $i \in {\cal V}^{(\lambda)}_\ell$ and $j \in  {\cal V}^{(\lambda)}_{\ell'}$ with $\ell \neq \ell'$; the choice 
$\widehat{\B{\Theta}}_{ij}=  \widehat{\M{W}}_{ij} =0$ 
satisfies the KKT conditions (\ref{kkt-1}) 
$$ | \s_{ij}  - \widehat{\M{W}}_{ij} | \leq \lambda$$
for all the off-diagonal entries in the block-matrix (\ref{matrix-block-diag1}).

By construction (\ref{sub-block-gl}) it is easy to see that for every $\ell$, 
the matrix $\widehat{\B{\Theta}}_{\ell}$ satisfies the KKT conditions (\ref{kkt-1}) and  (\ref{kkt-2}) corresponding to the 
$\ell^{\mathrm{th}}$ block of the $p \times p$ dimensional problem.
Hence $\widehat{\B{\Theta}}$ solves problem (\ref{eqn-1}).

The above argument shows that the connected components obtained from the estimated precision graph 
${\calG }^{(\lambda)}$  leads to a partition of the vertices $\{\widehat{\cal  V}^{(\lambda)}_\ell\}_{1 \leq \ell \leq \kaplam}$ such that for every $\ell \in \{1, \ldots, \bfk(\lambda)\}$, 
there is a $\ell' \in \{1,\ldots, \kaplam \}$ such that 
$\widehat{\cal  V}^{(\lambda)}_{\ell'} \subset {\cal  V}^{(\lambda)}_\ell$. In particular $\bfk(\lambda) \leq  \kaplam$.

Conversely, if $\widehat{\B{\Theta}}$ admits the decomposition as in the statement of the theorem, then 
it follows from (\ref{kkt-1}) that:\\
for $i \in \widehat{\cal V}^{(\lambda)}_\ell$ and $j \in  \widehat{\cal V}^{(\lambda)}_{\ell'}$ with $\ell \neq \ell'$;
$| \s_{ij}  - \widehat{\M{W}}_{ij} | \leq \lambda$.  
Since $\widehat{\M{W}}_{ij}=0$, we have $|\s_{ij} | \leq \lambda$.
This proves that the connected components of ${\bfG}^{(\lambda)}$ leads to a partition of the vertices, which is  
finer than the vertex-partition induced by the components of $\calG^{(\lambda)}$. In particular this implies that 
$\bfk(\lambda) \geq \kaplam  $.

Combining the above two we conclude $\bfk(\lambda) = \kaplam $ and also the  
equality (\ref{eq-conn-comp}). The permutation $\pi$ in the theorem appears since the labeling of the connected 
components is not unique. 
\end{proof}

\subsection{Proof of Theorem \ref{nested-comps-1}} \label{proof-thm2}
\begin{proof}
This proof is a direct consequence of Theorem \ref{thm-equal-1}, which establishes that 
the vertex-partitions induced by the 
the connected components of 
the estimated precision graph and the thresholded sample covariance graph are equal. 

Observe that, by construction, the connected components of the thresholded sample covariance graph i.e. $\bfG ^{(\lambda)}$ are nested within 
the connected components of $\bfG ^{(\lambda')}$. 
In particular, the vertex-partition induced by the components of the thresholded sample covariance graph 
at $\lambda$, is contained  
inside the vertex-partition induced by the components of the thresholded sample covariance graph at $\lambda'$.
Now, using Theorem \ref{thm-equal-1} we conclude that the vertex-partition induced by the components of the  estimated precision graph 
at $\lambda$, given by  $\{\widehat{\cal  V}^{(\lambda)}_\ell\}_{1 \leq \ell \leq \kaplam }$ is contained  
inside the vertex-partition induced by the components of the  estimated precision graph at $\lambda'$, given by $\{\widehat{\cal  V}^{(\lambda')}_\ell\}_{1 \leq \ell \leq \kaplamp }$.
The proof is thus complete. 
\end{proof}

\bibliographystyle{plainnat}
\bibliography{new_agst_new.bib}
\end{document}